\documentclass[10pt,twocolumn,letterpaper]{article}

\usepackage{iccv}
\usepackage{times}
\usepackage{epsfig}
\usepackage{subfig}
\usepackage{graphicx}
\usepackage{amsmath}
\usepackage{amssymb}
\usepackage{adjustbox}
\usepackage[table,xcdraw]{xcolor}

\usepackage[breaklinks=true,bookmarks=false]{hyperref}

\iccvfinalcopy 

\def\iccvPaperID{****} 

\ificcvfinal\fi

\begin{document}

\title{Object detection on aerial imagery using CenterNet}

\author{Dheeraj Reddy Pailla\\
International Institute of Information Technology\\
Hyderabad\\
{\tt\small dheerajreddy.p@students.iiit.ac.in}
\and
Varghese Kollerathu\\
Siemens Technology and Services Private Limited\\
Bengaluru\\
{\tt\small varghese.kollerathu@siemens.com}
\and
Sai Saketh Chennamsetty  \\
Siemens Technology and Services Private Limited\\
Bengaluru\\
{\tt\small sai.chennamsetty@siemens.com}
}

\maketitle
\ificcvfinal\thispagestyle{empty}\fi

\begin{abstract}
   Detection and classification of objects in aerial imagery have several applications like urban planning, crop surveillance, and traffic surveillance. However, due to the lower resolution of the objects and the effect of noise in aerial images, extracting distinguishing features for the objects is a challenge. We evaluate CenterNet, a state of the art method for real-time 2D object detection, on the VisDrone2019 dataset. We evaluate the performance of the model with different backbone networks in conjunction with varying resolutions during training and testing.
\end{abstract}

\section{Introduction}
\begin{figure*}
     \centering
     \subfloat[]{\includegraphics[width=0.33\textwidth]{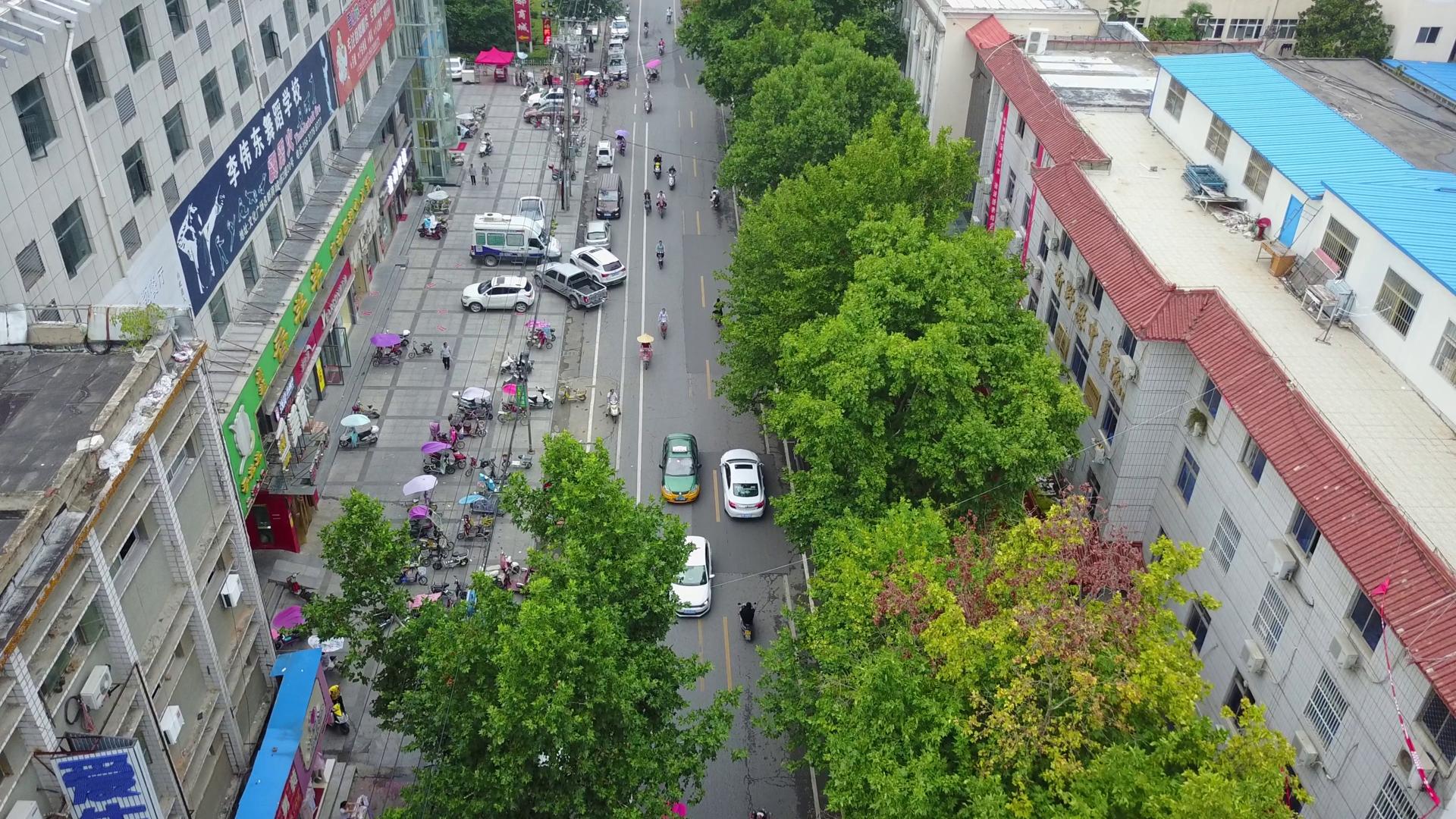}}\hfill
     \subfloat[]{\includegraphics[width=0.33\textwidth]{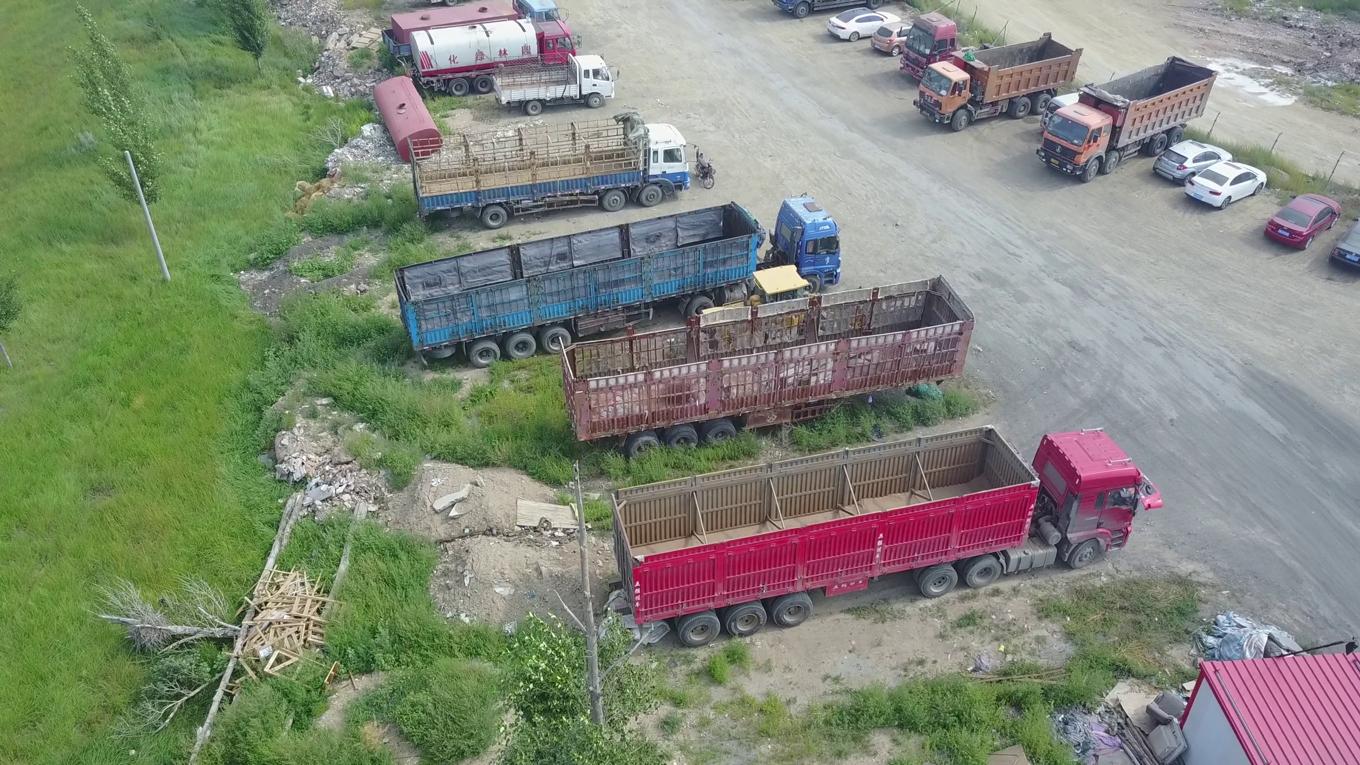}}\hfill
     \subfloat[]{\includegraphics[width=0.33\textwidth]{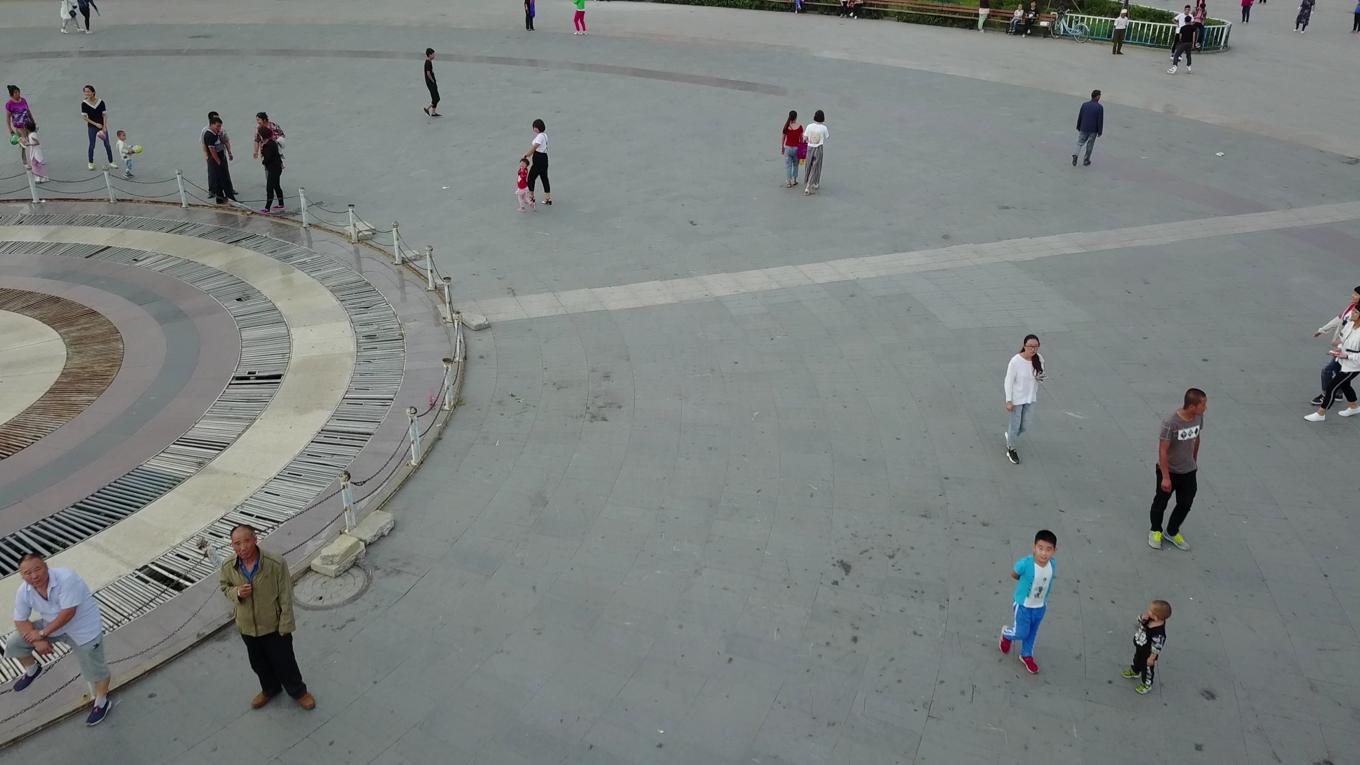}}\\
          \subfloat[]{\includegraphics[width=0.33\textwidth]{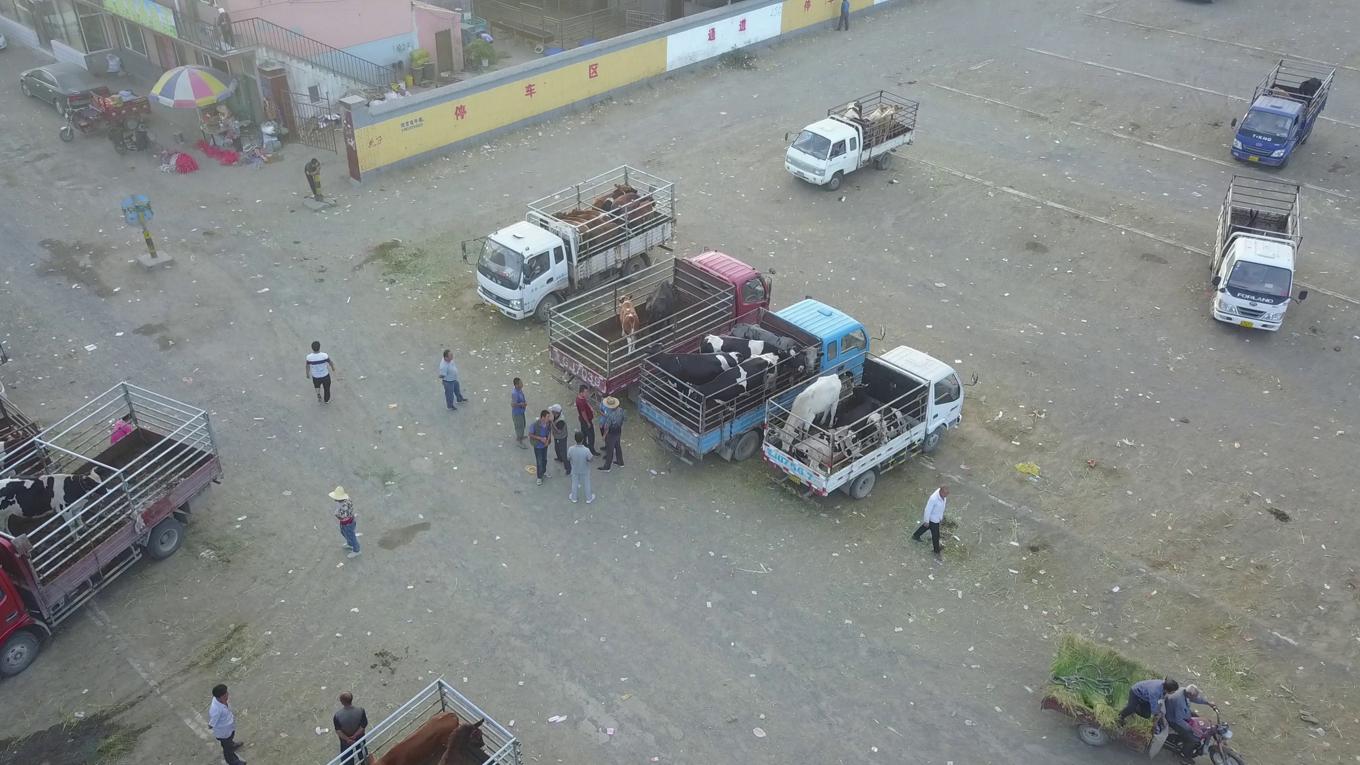}}\hfill
     \subfloat[]{\includegraphics[width=0.33\textwidth]{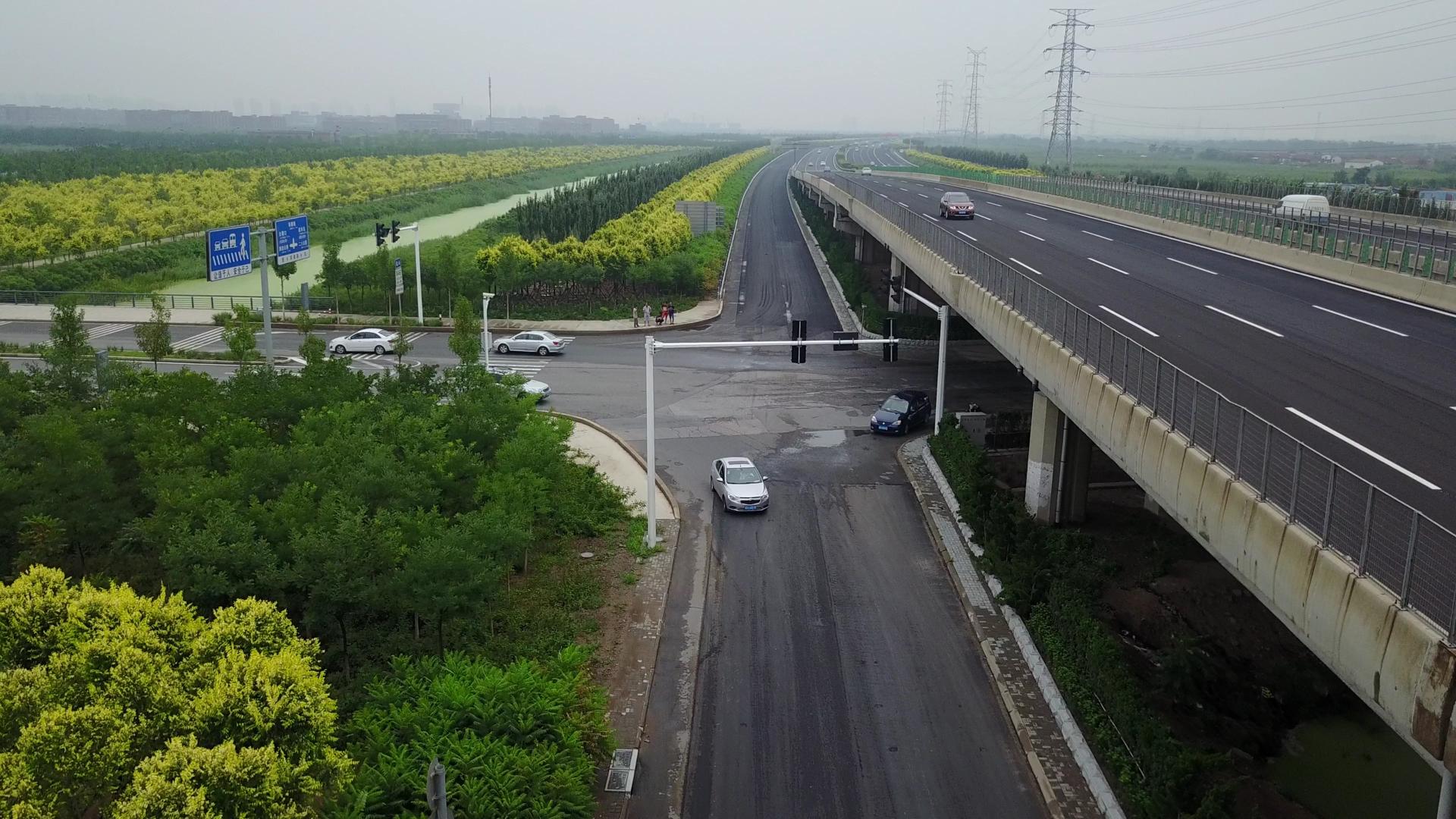}}\hfill
     \subfloat[]{\includegraphics[width=0.33\textwidth]{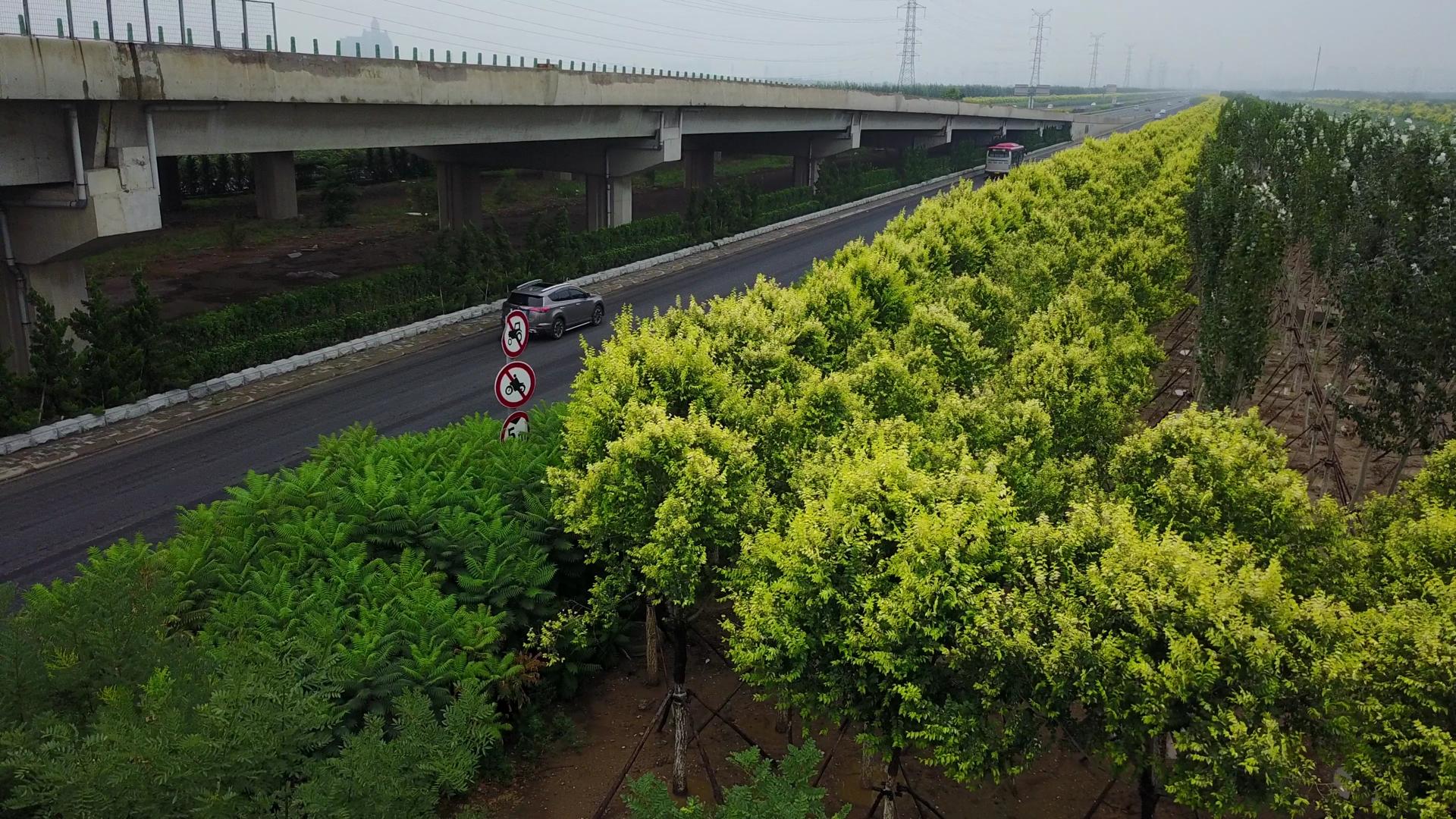}}\\
     \caption{Samples from the VisDrone2019 dataset.}
     \label{eady_state}
\end{figure*}

Images captured from drones differ from images of general objects captured using ordinary methods. Unusual aspect ratios, irregular points of view and lack of distinguishing details of objects in drone images are some of the differences between regular images and drone images. For example, MSCOCO \cite{lin2014microsoft} - a large-scale object detection, segmentation, and captioning dataset containing images of common objects taken in their general contexts. The foreground and the background are usually well separated and the images are of high resolution and quality. These object characteristics that help models in better identification are missing in drone images, as seen in Fig. \ref{eady_state}. 
Along with small object size, density of clustering of objects is high in drone footage.

\section{Related work}
In most scenarios the objects of interest occupies only a small region in the image. Object detection algorithms which are based on CNNs are broadly classified as two stage detectors and single stage detectors. Briefly, in the case of two stage detectors, the image is first passed through a pre-trained CNN to extract high level features. On the extracted feature maps, a fully convolutional network called Region Proposal Network (RPN) is applied to attain two outputs namely the probability if a region has an object or not and the co-ordinates of the bounding box. The region proposal network is trained so as to efficiently extract a predefined number (k=2000) regions from images. RPNs in two stage detectors decides if a region is a background or requires further processing. This enables them to achieve good generalization and performance.  However, this enhanced performance comes at the cost of requiring large inference time.

Single stage detectors overcome the challenges posed by RPNs  by using a fixed number of proposals, acquired through dense sampling of regions at different scales and aspect ratios.

\subsection{Two stage detectors}
R-CNN, one of the first successful detector, uses selective search to extract 2000 regions from the image, followed by a feature extractor and SVM to get the object scores and offset values. Fast-RCNN \cite{girshick2015fast}  generated region proposals on the feature map obtained by the convolution network instead of the input image. This made the pipeline faster as the convolution operation is only done once instead of 2000 times. Faster RCNN \cite{ren2015faster} further refines this work by replacing the selective search algorithm with a CNN.
\subsection{One stage detectors}
Presently, Single Shot MultiBox Detector(SSD)\cite{liu2016ssd} and YOLOv3\cite{redmon2018yolov3} are the most widely used one stage object detection models. SSD uses a modified VGG-16 model pretrained on ImageNet as its backbone with additional convolutional feature layers with progressively decreasing sizes. The Deconvolutional SSD (DSSD) improves on the SSD by use of deconvolution modules, which upsample the data and combine them with the feature layers of SSD. YOLO has a similar detection pipeline as SSD but trades off accuracy for speed by using a fixed grid cell aspect ratio and a lighter backbone. One-stage detectors, however, suffer from heavy imbalance between the foreground and background examples due to fixed sampling of candidate boxes. RetinaNet \cite {lin2017focal} tackles this problem by introducing focal loss, a variant of cross entropy loss that weighs down the loss assigned to well-classified examples. However, this still doesn't replace the heavy computation of fixed candidate boxes where a majority of them are part of the background. We evaluate a new detection network, CenterNet, which uses a keypoint estimation network to find potential objects, thus significantly reducing the inference time.

\section{Materials and Methods}
\subsection{Dataset}
We use the VisDrone2019 DET dataset for object detection in videos, and the VisDrone2019 VID dataset for object detection in videos. Both these datasets are prepared by the AISKYEYE team at the Lab of Machine Learning and Data Mining in Tianjin University, China.
\subsection{CenterNet}
Most successful object detectors, such as the aforementioned SSD  and YOLOv3, enumerate a nearly exhaustive list of potential object locations and classify each of them. This is wasteful, inefficient, and requires additional post-processing. CenterNet takes a different approach - it models an object as a single point - the center point of its bounding box. It uses keypoint estimation to find center points and regresses to all other object properties, such as size, 3D location, orientation, and even pose. This approach is end-to-end differentiable, simpler, faster, and more accurate than corresponding bounding box based detectors. 
We use CenterNet with an HourGlass-104 backbone \cite{zhou2019objects} for the task of detecting objects from aerial imagery. 
\subsubsection{Pre-processing}
The training and validation data provided by the AISKYEYE team were used to training and validating the models. The images were re-sized to 1024 $\times$ 1024 and normalized using ImageNet \cite{deng2009imagenet} statistics. 
\subsubsection{Training}
The model was initialized with weights pre-trained on the COCO \cite{lin2014microsoft} dataset. The learning rate was initialized at 2.5 e$^{-4}$ and was reduced by a factor of 10 at the end of the 90 and 120 epochs. The parameters of the network was optimized by using ADAM \cite{kingma2014adam} as the optimizer. 
\subsubsection{Testing}
During inference, the images were re-sized to 2048$\times$2048 and normalized using ImageNet statistics. We make use of 2 test time augmentation namely i) horizontal flip, ii) multi-scale testing at 0.5, 0.75, 1, 1.25 1.5 times the input resolution.   

\section{Results}
\par The performance of the evaluated on a variety of data-set viz; held out test data and challenge data.
\subsection{Evaluation Metric}
\par The performance of an model was evaluated on the basis of Average precision and average recall. The precision was measured at various IOU thresholds [0.5:0.05:0.95] by the bounding box generated by the algorithm and the ground truth.  The maximum number of detection (1,10,100 and 500) in the images were varied to compute average recall.

\subsection{Effect of backbone}
CenterNet provides the flexibility to use numerous models such as ResNet-18, DLA-34 and Hourglass-104. On the validation data (n=548 images), the performance of CenterNet  with each of the aforementioned networks as the backbone is given in Table \ref{backbone}. For each model, the images were re-sized to 512 $\times$ 512 and fed as input to each network. Additionally, we compare the performance of CentreNet with YOLOv3 (416x416). 

\begin{table}[]
\caption{Performance of CenterNet with different backbone on the validation data (n=548 images).}
\centering
\begin{tabular}{cc}
\hline
Backbone & mAP \\ \hline
ResNet18 & 13.36 \\ \hline
DLA-34 & 24.18 \\ \hline
Hourglass-104 & 31.97 \\ \hline
YOLOv3 (416x416) & 8 \\ \hline
\end{tabular}
\end{table}
\label{backbone}
\subsection{Effect of test time augmentation}
Augmentation of data during inference is often used technique in literature to minimize bias-variance. We compare the effect of having horizontal flip as a test-time augmentation scheme. Table \ref{augmentation} compares the performance of a Hourglass-104 CenterNet model with and without test-time augmentation. 
\begin{table}[]
\centering
\caption{Effect of test-time augmentation.}
\begin{tabular}{cc}
\hline
 Augmentation& mAP \\ \hline
No flip & 31.97 \\ \hline
Horizontal flip & 32.99 \\ \hline
\end{tabular}
\label{augmentation}
\end{table}
\par From the table, we observe that adding test-time augmentation improves the performance of the network by 1 \%. 

\subsection{Effect of multi-scale testing}
On images acquired from drones, various classes such as people, pedestrians to name few occupy fewer pixels than relatively larger objects such as cars, buses and trucks. Re-sizing images to 512 $\times$ 512 may lead to loss of discriminative features for smaller objects. In such scenarios, re-sizing images to larger dimension (say 1024 or 2048) is the often used technique. To further enhance the performance, we make use of multi-scale technique, wherein the images are scaled to different level such as 0.5, 0.75, 1, 1.25 \& 1.5.
\par We also study the association of the scaling parameter with respect to input resolution. From the experiments carried out, we observed that at 512x512 resolution using higher scales i.e. 1-2 produced higher performance than lower scales (0.5-1.5). A possible reason for this would be using lower scales would result in images which having even lower resolution than the input resolution (512x512).  At higher input resolution (2048x2048) scales from 0.5-1.5 produce better results than higher scales (1-2). From our experiments we observe that, increasing the resolution beyond 2048 by scaling (say 4096, scale =2) do not enhance the performance of the model. Apart from multi-scale testing, we also study the effect of horizontal flipping the images along with multi-scale testing. We observe that removing horizontal flip from the multi-scale testing during inference leads a dip in performance by approximately 2 \%.
\par
\begin{figure*}
     \centering
     \subfloat[]{\includegraphics[width=0.50\textwidth]{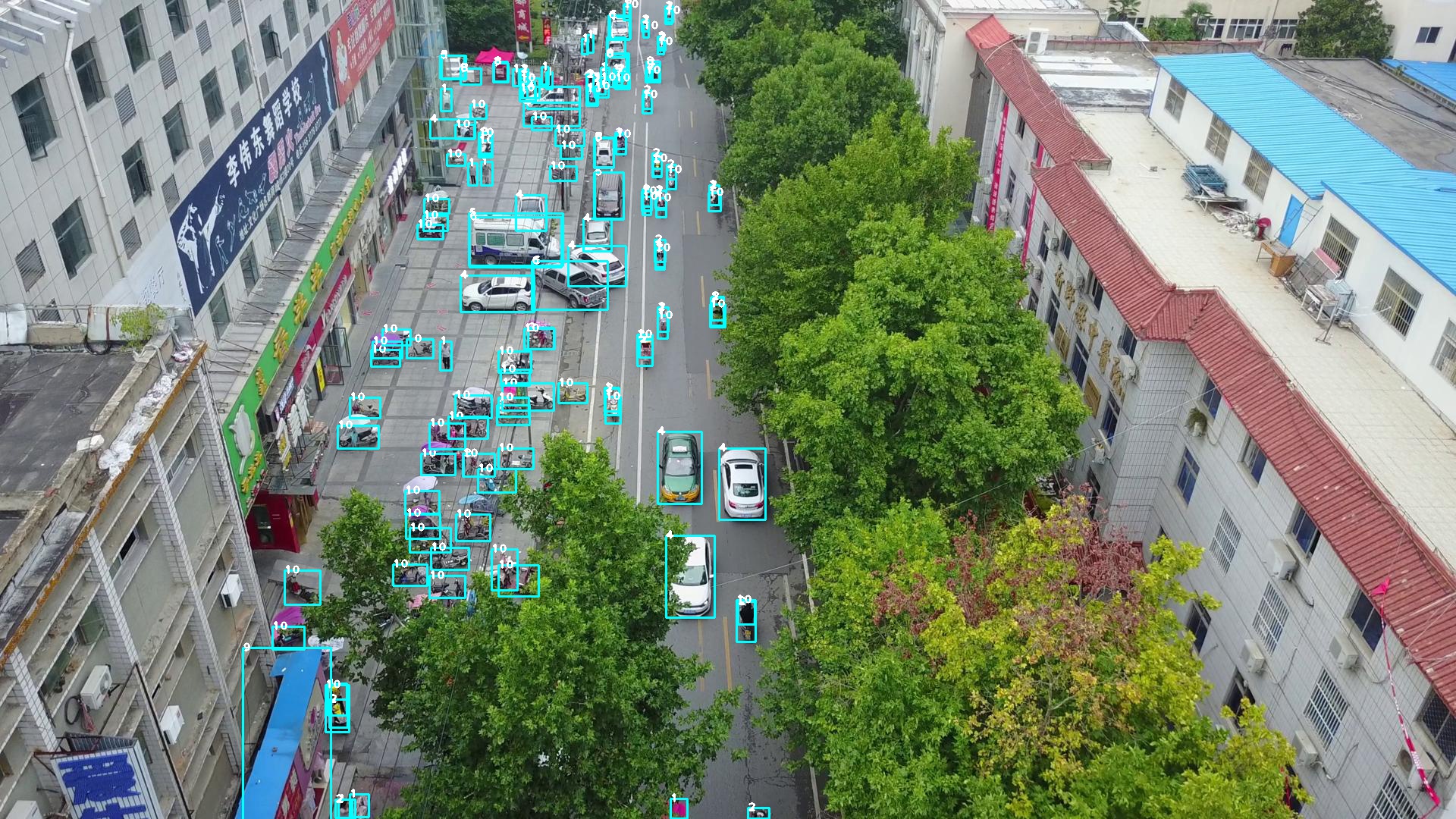}}\hfill
     \subfloat[]{\includegraphics[width=0.50\textwidth]{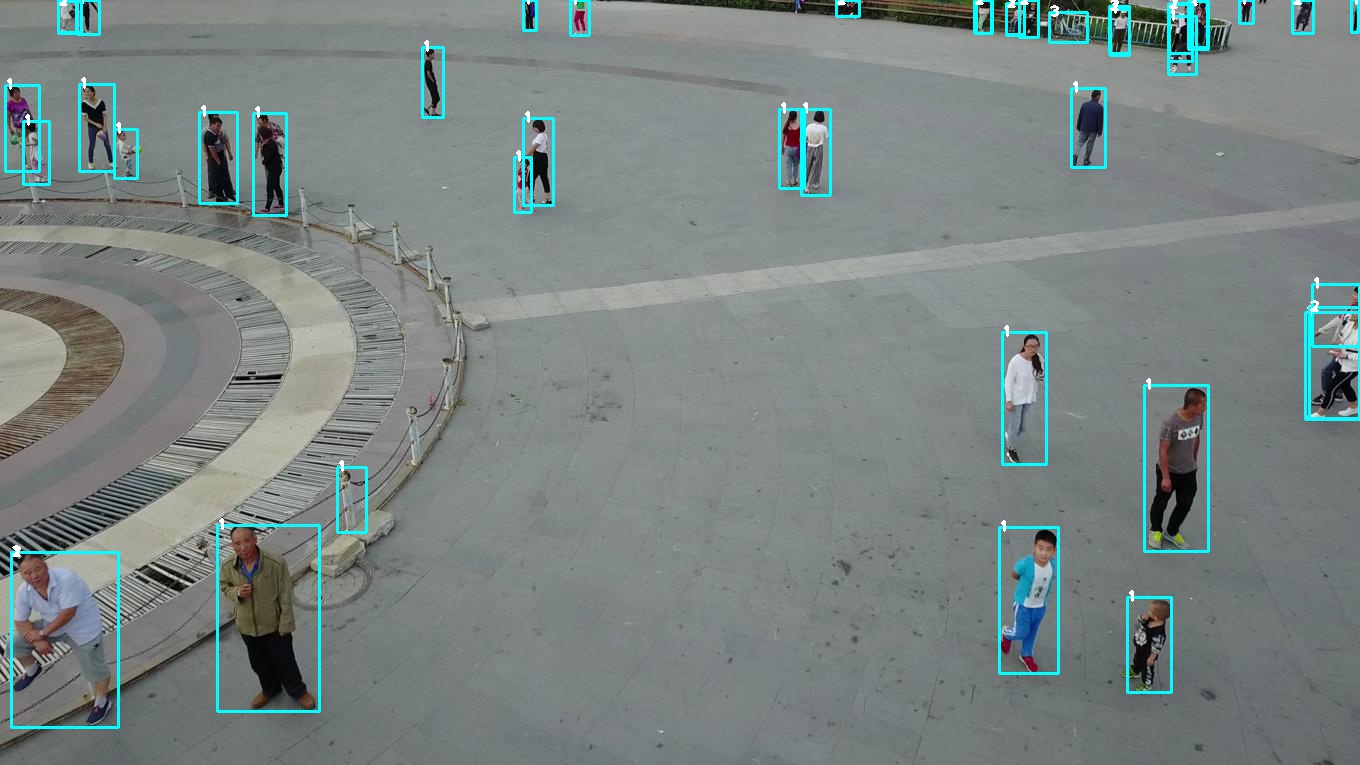}}\\
     \subfloat[]{\includegraphics[width=0.60\textwidth]{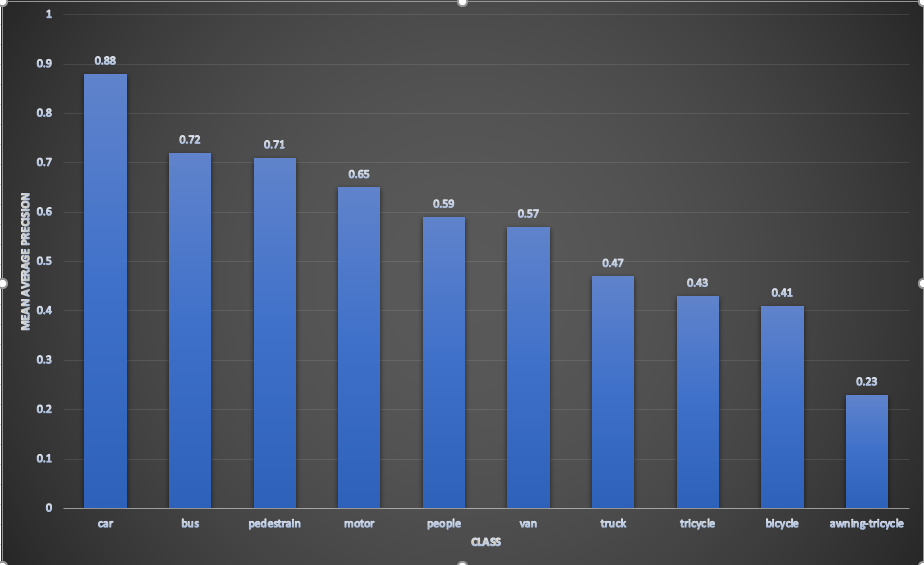}}\\

     \caption{Performance of model on the validation data (n=528). a) Class-wise performance on the validation data. b) Performance of the model on sample data from validation data (0000001-02999), c) Performance on validation data (0000069-01878). }
     \label{steady_state}
\end{figure*}

 Based on the performance, we set the input resolution of the image to 2048 and the scale to be 0.5, 0.75, 1, 1.25 and 1.5. The performance of the model across each class in given in Fig. \ref{steady_state}.

\begin{table}[]
\centering
\caption{Effect of input resolution. At each resolution, we infer at various scales.For each range of scale, the step size associated to scaling parameter is set to 0.25. Additionally at each scale we also include test-time augmentation (horizontal flip).}
\begin{tabular}{ccc}
\hline
 Input Resolution &Scale & mAP \\ \hline
512 & 0.5-1.5& 43.17 \\ \hline
512 & 1-2.5& 49.10 \\ \hline
2048 & 0.5-1.5 & 58.03\\ \hline
2048 & 1-2 & 51.99\\ \hline
2048 (no-flip) & 0.5-1.5 & 56.88 \\ \hline
\end{tabular}
\label{augmentation}
\end{table}

\subsection {Performance on Challenge data}
The performance of all the competitors were evaluated on the data-set provided by the challenge organizers. For the task of detecting objects from images (Track 1), the data-set comprises of 3,190 images. On the test data, our algorithm stands 7$^{th}$ on the leader-board with an overall mAP of 27.83 \%. Table \ref{track1_metricwise}, illustrates performance of the algorithm across various classes. Table \ref{track1_metricwise}, compares the performance of the proposed solution against other top performing algorithms.  
\par Additionally, we also participate in Track 2 of the challenge, i.e. detection of objects on videos. The dataset comprises of 33 sequences (12,968 images). The same model used in Track 1 without any sort of re-training or fine-tuning was utilized for Track 2. On the leader-board, the proposed algorithm was placed 5$^{th}$ and Table \ref{track2_metricwise} compares the performance of the solution with other top performing algorithms. Table \ref{track1andtrack2} illustrates the class-wise performance of the solution for Track1 and Track2 of the Visdrone-2019 dataset.

\begin{table*}[]
\caption{Comparison of the proposed solution against top performing techniques on the Track1 challenge data (n=3190) image. In the table, an entry in \textcolor{red}{red} depicts the best performance in that particular when compared other competing algorithms.}
\centering
\begin{tabular}{ccccccccc}
\hline
Position & Method & AP{[}\%{]} & AP50{[}\%{]} & AP75{[}\%{]} & AR1{[}\%{]} & AR10{[}\%{]} & AR100{[}\%{]} & AR500{[}\%{]} \\ \hline
1 & DPNet-ensemble & {\color[HTML]{FE0000} 29.62} & 54 & {\color[HTML]{FE0000} 28.7} & 0.58 & 3.69 & 17.1 & 42.37 \\ \hline
2 & RRNet & 29.13 & {\color[HTML]{FE0000} 55.82} & 27.23 & 1.02 & {\color[HTML]{FE0000} 8.5} & {\color[HTML]{FE0000} 35.19} & 46.05 \\ \hline
3 & ACM-OD & 29.13 & 54.07 & 27.38 & 0.32 & 1.48 & 9.46 & 44.53 \\ \hline
4 & S+D & 28.59 & 50.97 & 28.29 & 0.5 & 3.38 & 15.95 & 42.72 \\ \hline
5 & BetterFPN & 28.55 & 53.63 & 26.68 & 0.86 & 7.56 & 33.81 & 44.02 \\ \hline
6 & HRDet & 28.39 & 54.53 & 26.06 & 0.11 & 0.94 & 12.95 & 43.34 \\ \hline
7 & \textbf{CN-DhVaSa(ours)} & 27.83 & 50.73 & 26.77 & 0 & 0.18 & 7.78 & {\color[HTML]{FE0000} 46.81} \\ \hline
20 & TridentNet & 22.51 & 43.29 & 20.5 & {\color[HTML]{FE0000} 1.17} & 8.3 & 28.98 & 39.84 \\ \hline
\end{tabular}
\label{track1_metricwise}
\end{table*}
\begin{table*}[]
\caption{Comparison of the proposed solution against top performing techniques on the Track2 challenge data (n=33) sequeces. In the table, an entry in \textcolor{red}{red} depicts the best performance in that particular when compared other competing algorithms.}
\centering
\begin{tabular}{lllllllll}
\hline
Position & Method & AP{[}\%{]} & AP50{[}\%{]} & AP75{[}\%{]} & AR1{[}\%{]} & AR10{[}\%{]} & AR100{[}\%{]} & AR500{[}\%{]} \\ \hline
1 & DBAI-Det & {\color[HTML]{FE0000} 29.22} & {\color[HTML]{FE0000} 58} & {\color[HTML]{FE0000} 25.34} & {\color[HTML]{FE0000} 14.3} & {\color[HTML]{FE0000} 35.58} & {\color[HTML]{FE0000} 50.75} & {\color[HTML]{FE0000} 53.67} \\ \hline
2 & AFSRNet & 24.77 & 52.52 & 19.38 & 12.33 & 33.14 & 45.14 & 45.69 \\ \hline
3 & HRDet+ & 23.03 & 51.79 & 16.83 & 4.75 & 20.49 & 38.99 & 40.37 \\ \hline
4 & VCL-CRCNN & 21.61 & 43.88 & 18.32 & 10.42 & 25.94 & 33.45 & 33.45 \\ \hline
5 & \textbf{CN-DhVaSa(ours)} & 21.58 & 48.09 & 16.76 & 12.04 & 29.6 & 39.63 & 40.42 \\ \hline
\end{tabular}
\label{track2_metricwise}
\end{table*}

\begin{table*}[]
\caption{Category wise performance of top performing algorithms on Track 1 and Track 2 challenge data.}
\centering
\begin{tabular}{ccccccccccc}
\hline
Input & ped & people & bicycle & car & van & truck & tricycle & awn & bus & motor \\ \hline
Image & 31.05 & 12.99 & 9.08 & 51.92 & 38.33 & 31.14 & 24.24 & 21.06 & 40.94 & 20.35 \\ \hline
Video & 27.86 & 6.59 & 12.47 & 33.92 & 29.91 & 40.55 & 13.99 & 12.91 & 24.48 & 6.98 \\ \hline
\end{tabular}
\label{track1andtrack2}
\end{table*}
\
\section{Conclusion}
Typically, one stage detectors such as YOLOv3 and SSD don't perform particularly well in instances of small object detection - specifically detection in aerial imagery. In this paper, we make use of Centernet for detection of objects  from images and videos. 
\par We evaluate various backbone networks such as ResNet-18, DLA-34 and Hourglass-104. From the experiments, we observe that the HourGlass-104 backbone produced the best performance when compared to other networks and a standard YOLOv3.

\par We observe that test-time augmentation such as horizontal flipping of image and multi-scale testing aid in enhancing the overall performance of the model. 
\par On the the challenge data provided by the organizers, the solution attained 7$^{th}$ position for the task of detecting objects from images.  When compared to other techniques, the solution was attained best average recall with 500 detection per image. 
Without any fine-tuning or re-training, the model used for detecting objects from images was used to Track 2 (detection of objects from videos). On the leader-board for object detection from videos, the solution achieved a competitive 5$^{th}$ position. 



{\small
\bibliographystyle{ieee}
\bibliography{main}
}

\end{document}